\title{Poisson Inverse Problems by the Plug-and-Play scheme}
\author{
        Arie Rond, Raja Giryes and Michael Elad\\
        Department of Computer Science Technion---Israel Institute of Technology\\
        Technion City, Haifa 32000, Israel\\
            \texttt{sarikr@cs.technion.ac.il}
        \textbar{}
            \texttt{raja.giryes@duke.edu}
        \textbar{}
	        \texttt{elad@cs.technion.ac.il}
	        \footnote
	        {
	        	This research was supported by the European Research Council under EU’s 7th Framework Program, ERC grant agreement 320649, by the Intel Collaborative Research Institute for Computational Intelligence, and by the Google Faculty Research Award.
	        }
}
\date{\today}
\documentclass[11pt]{article}

\usepackage[paper=a4paper,dvips,top=1.5cm,left=1.5cm,right=1.5cm,foot=1cm,bottom=1.5cm]{geometry}
\usepackage{times}
\usepackage{booktabs}
\usepackage{amsmath}
\usepackage{algpseudocode}
\usepackage{algorithm}
\usepackage{pifont}
\usepackage{graphicx}
\usepackage{epstopdf}
\usepackage{paralist}
\usepackage{float}
\usepackage{hhline}
\usepackage{caption}
\usepackage{subcaption}
\usepackage[titletoc,title]{appendix}
\restylefloat{table}
\graphicspath{{./results/}}
\begin{document}
\maketitle

\begin{abstract}
The Anscombe transform \cite{anscombe1948transformation} offers an approximate conversion of a Poisson random variable into a Gaussian one. This transform is important and appealing, as it is easy to compute, and becomes handy in various inverse problems with Poisson noise contamination. Solution to such problems can be done by first applying the Anscombe transform, then applying a Gaussian-noise-oriented restoration algorithm of choice, and finally applying an inverse Anscombe transform. The appeal in this approach is due to the abundance of high-performance restoration algorithms designed for white additive Gaussian noise (we will refer to these hereafter as "Gaussian-solvers"). This process is known to work well for high SNR images, where the Anscombe transform provides a rather accurate approximation. When the noise level is high, the above path loses much of its effectiveness, and the common practice is to replace it with a direct treatment of the Poisson distribution. Naturally, with this we lose the ability to leverage on vastly available Gaussian-solvers.

In this work we suggest a novel method for coupling Gaussian denoising algorithms to Poisson noisy inverse problems, which is based on a general approach termed "Plug-and-Play"  \cite{venkatakrishnan2013plug}. Deploying the Plug-and-Play approach to such problems leads to an iterative scheme that repeats several key steps:
\begin{inparaenum}[(i)]
	\item
	A convex programming task of simple form that can be easily treated;
	\item
	A powerful Gaussian denoising
	algorithm of choice; and
	\item
	A simple update step.
\end{inparaenum}
Such a modular method, just like the Anscombe transform, enables other developers to plug their own Gaussian denoising algorithms to our scheme in an easy way. While the proposed method bares some similarity to the Anscombe operation, it is in fact  based on a different mathematical basis, which holds true for all SNR ranges.
\end{abstract}

\section{Introduction}
	In an inverse problem we are given a degraded image, $y$, and want to recover from it a clean image, $x$. The mathematical relation between the two images is given by $y=\mathcal{N}\left(Hx\right)$, where $H$ is some linear operator and $\mathcal{N}$ is a noise operator. A popular way to handle this reconstruction is to use a Bayesian probabilistic model that contains two ingredients:
	\begin{inparaenum}[(i)]
	\item
		the measurement forward model, mathematically given by $P\left(y|x\right)$ ; and
	\item
		a prior model for clean images, given by $P\left(x\right)$
\end{inparaenum}
.

In our work we concentrate on the case of Poisson Inverse Problems (PIP), Where $\mathcal{N}$ stands for Poisson contamination. In a Poisson model for an image the gray levels of the image pixels are viewed as Poisson distributed random variables. More specifically, given a clean image pixel $x[i]$, the probability of getting a noisy value $y[i]$ is given by

 \begin{equation}
	 P\left( {y\left[ i \right]|x\left[ i \right]} \right) =
	 \begin{cases}
		 \frac{{{{\left( {x\left[ i \right]} \right)}^{y\left[ i \right]}}}}{{y\left[ i \right]!}}{e^{ - x\left[ i \right]}} &\mbox{if } x\left[ i \right] > 0 \\
		 
		 \delta \left( {y\left[ i \right]} \right) &\mbox{if } x\left[ i \right] = 0
 	\end{cases}
 	.
 \end{equation}

A known property of this distribution  is that $x\left[i\right]$ is both the mean and variance of $y\left[i\right]$. This model is relevant in various tasks such as very low light imaging, CT reconstruction \cite{rodrigues2008denoising}, fluorescence microscopy \cite{boulanger2010patch}, astrophysics \cite{schmitt2010poisson} and spectral imaging \cite{keenan2004accounting}. Common to all these tasks is the weak measured signal intensity.

An important note about Poisson noise is that the SNR of the measurements is proportional to the original image intensity, given by $\sqrt{x[i]}$. Therefore the peak value of an image is an important characteristic, needed when evaluating the level of noise in the image. For high peak levels, there exist several very effective ways to solve Poisson inverse problems. Many of these methods rely on the fact that it is possible to perform an approximate transform (known as Variance Stabilized Transform - VST) of the Poisson distribution into a Gaussian one \cite{anscombe1948transformation}, \cite{fisz1955limiting}. Since there are highly effective algorithms for Gaussian noise restoration (e.g. \cite{dabov2007image}, \cite{elad2006image}, \cite{mairal2009non},  \cite{yu2012solving}, \cite{danielyan2012bm3d}), such methods can be used, followed by an inversion of the VST operation after the Gaussian solver \cite{makitalo2011optimal}, \cite{zhang2008wavelets}. 

When dealing with lower peaks, such transformations become less efficient, and alternative methods are required, which treat the Poisson noise directly (e.g. \cite{giryes2014sparsity}, \cite{salmon2014poisson}). In recent years this direct approach has drawn considerable attention, and it seems to be very successful. In this work we aim at studying yet another method for Poisson inverse problem restoration that belongs to the direct approach family. The appeal of the proposed method is the fact that it offers an elegant bridge between the two families of methods, as it is relying too on Gaussian noise removal, applied iteratively.

This paper is organized in the following way : In section \ref{section_P&P} we introduce the plug and play approach, as presented in \cite{venkatakrishnan2013plug}. We also extend this scheme to be able to work with several priors in parallel. In section \ref{section_ALGORITHM} we present our algorithm, as derived from the plug and play approach. This section explains how to integrate a custom Gaussian denoising algorithm of choice,and discusses several improvements that were added to the algorithm. In section \ref{section_EXPERIMENTS} we present experiments results, and in section \ref{section_DISCUSSION} we conclude our paper by suggesting further improvements.

\section{The Plug-and-Play (PaP) Approach}\label{section_P&P}
\subsection{Standard Plug-and-Play}
The Plug-and-Play framework, proposed by Venkatakrishnan, Bouman and Wohlberg \cite{venkatakrishnan2013plug}, allows simple integration between inversion problems and priors, by applying a Gaussian denoising algorithm, which corresponds to the used prior. One of the prime benefits in the PaP scheme is the fact that the prior to be used does not 
have to be explicitly formulated as a penalty expression. Instead, the idea is to split the prior from the inverse problem, a task that is done elegantly by the ADMM optimization method \cite{boyd2011distributed}, and then the prior is deployed indirectly by activating a Gaussian denoising algorithm of choice.

The goal of the PaP framework is to maximize the posterior probability in an attempt to implement the MAP estimator. Mathematically, this translate to the following:
\begin{equation}
	\max \limits_{x \in R^{m \times n}} P\left( x|y \right)=
	\max \limits_{x \in R^{m \times n}} \frac{{P\left( {y|x} \right)P\left( x \right)}}{{P\left( y \right)}}=
	\max \limits_{x \in R^{m \times n}} P\left( y|x \right)P\left( x \right).
\end{equation}
The above suggests to maximize the posterior probability $P(x|y)$ with respect to the ideal image $x$, which is of size $n \times m$ pixels. Taking element wise $-ln\left( \cdot \right)$ of this expression gives an equivalent problem of the form
\begin{equation}
\min \limits_{x \in R^{m \times n}} -ln\left(P\left( x|y \right)\right)=
\min \limits_{x \in R^{m \times n}} -ln\left(P\left( y|x \right)\right)-ln\left(P\left( x \right)\right).
\end{equation}
In order to be consistent with \cite{venkatakrishnan2013plug} we denote
$l\left(x\right)= -ln\left(P\left( y|x \right)\right)$
and
$s\left(x\right)=-ln\left(P\left( x \right)\right)$.
Thus our task is to find $x$ that solves the problem 
\begin{equation}
\hat x = \mathop {\arg\min}\limits_{x \in {R^{m \times n}}} l\left( x \right) + \beta s\left( x \right).
\end{equation}
Note that $y$ is constant in this minimization. Also, a parameter $\beta$ was added to achieve more flexibility.
By adding a variable splitting technique to the optimization problem we get
\begin{equation} \label{eq:VAR_SPLITTING}
\begin{aligned}
\hat x = &\mathop {\arg\min }\limits_{x,v \in {R^{m \times n}}} l\left( x \right) + \beta s\left( v \right). \\
&s.t\,\,\,x = v
\end{aligned}
\end{equation}
This problem can be solved using ADMM \cite{boyd2011distributed} by constructing an augmented Lagrangian which is given by
\begin{equation}
{L_\lambda } = l\left( x \right) + \beta s\left( v \right) + \frac{\lambda }{2}\left\| {x - v + u} \right\|_2^2 - \frac{\lambda }{2}\left\| u \right\|_2^2.
\end{equation}
ADMM theory \cite{boyd2011distributed} states that minimizing \eqref{eq:VAR_SPLITTING} is equivalent to iterating until convergance  over the following three steps:

\begin{equation}
	\begin{array}{l}
	{{x}^{k + 1}} = \mathop {\arg \min }\limits_x {L_\lambda }\left( {x,{v^k},{u^k}} \right),\\
	{{v}^{k + 1}} = \mathop {\arg \min }\limits_v {L_\lambda }\left( {{{x}^{k + 1}},v,{u^k}} \right),\\
	{u^{k + 1}} = {u^k} + \left( {{{x}^{k + 1}} - {{v}^{k + 1}}} \right).
	\end{array}
\end{equation}

By plugging $L_\lambda$ we get 
\begin{equation}
	\begin{array}{l}
	{{x}^{k + 1}} = \mathop {\arg \min }\limits_x l\left( {y|x} \right) + \frac{\lambda }{2}\left\| {x - \left( {{v^k} - {u^k}} \right)} \right\|_2^2,\\
	{{v}^{k + 1}} = \mathop {\arg \min }\limits_v \frac{\lambda }{2}\left\| {{x^{k + 1}} + {u^k} - v} \right\|_2^2 + \beta s\left( v \right),\\
	{u^{k + 1}} = {u^k} + \left( {{{x}^{k + 1}} - {{v}^{k + 1}}} \right).
	\label{eq::PaP steps}
	\end{array}
\end{equation}
The second step means applying a Gaussian denoising algorithm which assumes a prior $s\left(v\right)$ on the image ${x^{k + 1}} + {u^k}$ with variance of $\sigma^2=\frac{\beta}{\lambda}$. Therefore, as already mentioned above, we do not have to know the formulation of the prior explicitly, as we can simply use a Gaussian denoising algorithm which corresponds to it.

The first step is dependent on the inversion problem we are trying to solve. In the next section we show how Poisson inverse problems are connected to this step. We will see that in this case step 1 is convex and becomes easy to compute. When handling the Poisson Denoising problem, this steps becomes even simpler because it is also separable, thus leading to a scalar formula that resembles the Anscombe transform.

\subsection{Plug-and-Play Extension}
We now show a simple extension of the Plug-and-Play method that enables to use several Gaussian denoisers.
We start from the following ADMM formulation, that follows Equation (\ref{eq:VAR_SPLITTING})

\begin{equation}
\begin{aligned}
&{\arg\min } \,\, l(x)+\beta_1 s_1(v_1)+\beta_2 s_2(v_2),
\\
&s.t\,\,x=v_1,x=v_2
\end{aligned}
\end{equation} 
where $s_1$ and $s_2$ are two priors that we aim to use, and $v_1$ and $v_2$ are two auxiliary variables that will help in simplifying the solution of this problem. Following the steps taken above in the derivation of the PaP, we get

\begin{enumerate}
	\item [Step 1:]
		\begin{equation}
			{x^{k + 1}} = \mathop {\arg \min }\limits_x l\left( x \right) + \lambda \left\| {x - {v_1}^k + {u_1}^k} \right\|_2^2 + \lambda \left\| {x - {v_2}^k + {u_2}^k} \right\|_2^2.
		\end{equation}
		As in the original Plug-and-Play, this expression too is convex if $l(x)$ is convex, and also separable if $l(x)$ is separable.
		
	\item [Step 2:]		
		\begin{equation}
		\begin{array}{l}
		{v_1}^k = \mathop {\arg \min }\limits_v {{\beta }_1}{s_1}\left( {{v_1}} \right) + \lambda \left\| {{x^{k + 1}} - {v_1} + {u_1}^k} \right\|_2^2\\
		{v_2}^k = \mathop {\arg \min }\limits_v {{\beta }_2}{s_2}\left( {{v_2}} \right) + \lambda \left\| {{x^{k + 1}} - {v_2} + {u_2}^k} \right\|_2^2
		\end{array}
		\end{equation}
		which are two Gaussian denoising steps, each using a different prior.
	\item [Step 3:]
	
			\begin{equation}
			\begin{array}{l}
				{u_1}^{k + 1} = u_1^k + {x^{k + 1}} - {v_1}^{k + 1}\\
				{u_2}^{k + 1} = u_2^k + {x^{k + 1}} - {v_2}^{k + 1}
			\end{array}
			\end{equation}	
\end{enumerate}
Obviously, this scheme can be generalized to use as many priors as needed. The core idea behind this generalization is that often times we may encounter different priors that address different features of the unknown image, such as self-similarity, local smoothness or other structure, scale-invariance, and more. By merging two such priors into the PaP scheme, we may get an overall benefit, as they complement each other. 

\section{P$^{\boldsymbol{4}}$IP Algorithm}\label{section_ALGORITHM}
We now turn to introduce the "Plug-and-Play Prior for Poisson Inverse Problem" algorithm, P$^4$IP in short, and how it uses the plug and play framework. We start by invoking the proper log-likelihood function $l(x)$ into the above-described formulation, this way enabling the integration of Gaussian denoising algorithms to the Poisson inverse problems. Then we discuss two applications of our algorithm -- the denoising and the deblurring scenarios.
 
\subsection{The Proposed Algorithm}
We denote an original (clean) image, with dimensions $m\times n$, by an $m \times n$  column-stacked vector $x$. Similarly, we denote a noisy image by $y$. The $i$'s pixel in $x$ (and respectively $y$) is given by $x[i]$ (respectively $y[i]$). We also denote by $H$ the linear degradation operator that is applied on the image, which could be a blur operator, down-scaling or even a tomographic projection.
In order to proceed we should find an expression for $l(x)$. As mentioned before, this is given by taking $-ln\left( \cdot \right)$ of $P(y|x)$. When taking $H$ into account we get
\begin{equation}
P\left( {y|x} \right) = \prod\limits_i {\frac{{{{\left( {Hx} \right)}_i}^{{y_i}}}}{{\Gamma \left( {{y_i} + 1} \right)}}{e^{ - {{\left( {Hx} \right)}_i}}}} .
\end{equation}
Thus, $l(x)$ is given by
\begin{equation}
l(x)=\ln(P(y|x))=\sum\limits_i {\ln \left( {\frac{{{{\left( {Hx} \right)}_i}^{{y_i}}}}{{\Gamma \left( {{y_i} + 1} \right)}}{e^{ - {{\left( {Hx} \right)}_i}}}} \right)}=-y^T ln(Hx)+1^THx+constant.
\end{equation}
Relying on equation (\ref{eq::PaP steps}), the first ADMM step in matrix form is therefore 
\begin{equation}
	\mathop {\arg \min }\limits_x {L_\lambda } = \mathop {\arg \min }\limits_x  - {y^T}\ln \left( {Hx} \right) + {1^T}Hx + \frac{\lambda }{2}\left\| {x - v + u} \right\|^2_2.
\end{equation}
This expression is convex and can be solved quite efficiently by modern optimization methods. The final algorithm is shown in Algorithm~\ref{Inverse problem algorithm}.

\begin{algorithm}
	\caption{- P$^{\boldsymbol{4}}$IP}
	\label{Inverse problem algorithm}
	\begin{algorithmic}
		\State \textbf{Input}: Distorted image $y$, Gaussian\_denoise$\left(\cdot \right)$ function
		
		\State Initialization: set $k=0$,$u^{0}=0,v^{0}= $some initialization;
		\While {!stopping criteria}	
		\State
		$
		{{x}^{k + 1}} = \mathop {\arg \min }\limits_x  - {y^T}\ln \left( {Hx} \right) + {1^T}Hx + \frac{\lambda }{2}\left\| {x - v^{k} + u^k} \right\|_2^2
		$
		\State	
		$v^{k + 1}=$ Gaussian\_denoise $({x^{k + 1}} + {u^k})$ with $ \sigma^2=\frac{\beta}{\lambda}$
		\State
		$
		{u^{k + 1}} = {u^k} + \left( {{{x}^{k + 1}} - {{v}^{k + 1}}} \right)
		$	
		\State
		$k=k+1$
		\EndWhile
		\State \textbf{Output}: Reconstructed image $x^k$
	\end{algorithmic}
\end{algorithm}

 Obviously we could use the Plug-and-Play extension that employs several denoising methods, as shown in the previous section. Such a change requires only slight modifications to Algorithm~\ref{Inverse problem algorithm}.
 
\subsubsection{Poisson Denoising}
\begin{figure}[H]
	\centering

		\includegraphics[scale=0.8,clip,trim=0cm 0cm 0cm 0.7cm]{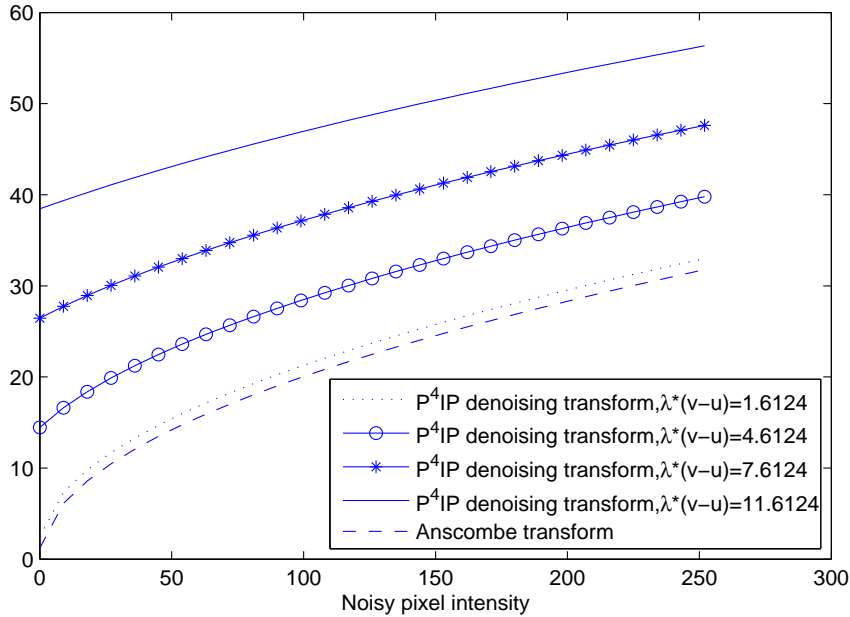}
		\label{img::p4ip_ansc_transforms}
	
	\caption{P$^4$IP and Anscombe transformations as a function of input noisy pixel. $\lambda=0.25$.}
	\label{plot::denoising transforms}
\end{figure}

For the special case of Poisson denoising $H=I$. In this case the first ADMM step is separable, which means that it could be solved for each pixel individually. Moreover, this step can be solved by the closed form solution
\begin{equation}
x^{k + 1}[i] = \frac{{\left( {\lambda \left( {v^k[i] - u^k[i]} \right) - 1} \right) + \sqrt {{{\left( {\lambda \left( {v^k[i] - u^k[i]} \right) - 1} \right)}^2} + 4\lambda {y[i]}} }}{{2\lambda }},
\label{eq::denoising step 1 equation}
\end{equation}
where $x^k[i]$ is the i'th pixel of $x^k$ (and $v^k[i]$, $u^k[i]$ and $y[i]$ are the i'th pixels of $v^k$, $u^k$ and $y$ respectively). The full derivation of this step is shown in the appendix \ref{appendix_DENOISING_ADMM_FIRST_STEP}.
A closer look at this expression reveals some resemblance to the Anscombe transform. Indeed, for the initial condition
$u^0=0, v^0=4\left(\sqrt{\frac{3}{8}}+1\right)$, and $\lambda=0.25$,
the variance of $x$ is the same as the one achieved by Anscombe's transform, because they differ only by a constant.
We mention here that we did not find that the initialization of the parameters lead to noticeable change in the final reconstruction, as long as it is the same order of magnitude of the noisy image, and therefore, all the shown results use all zero image as initialization.
Figure \ref{plot::denoising transforms} shows the transform done by Equation (\ref{eq::denoising step 1 equation}) for $\lambda =0.25, v^k[i]-u^k[i]=4\left(\sqrt{\frac{3}{8}}+1\right)+i$ for $i=\{ 0, 3, 6, 9 \}$, and the Anscombe transform. While this curve may look like the Anscombe one, PaP is substantially different in two ways - 
\begin{inparaenum}[(i)]
	\item
	this curve changes (locally) from one iteration to another due to the change in $u$ and $v$, and
	\item
	 we do not apply the inverse transform after the Gaussian denoising. 
 \end{inparaenum}
 
\subsubsection{Poisson Deblurring}
When dealing with the deblurring problem, $H$ represents a blur matrix. The first ADMM step is no longer separable and usually no analytical solution is available. However the problem is convex and a common way to solve it is by using iterative optimization methods, which usually require the gradient. Turning back to our problem, the gradient of $L_\lambda(x)$ is given by 
\begin{equation}
{\nabla _x}{L_\lambda } =  - {H^T}\left(y/\left(Hx\right)\right) + {H^T}1 + \lambda \left( {x - v + u} \right).
\end{equation}
Where "/" stands for element-wise division. As can be seen, this gradient is easy to compute, as it requires blurring the temporary solution $x$, the constant vector $1$ and the vector $y/\left(Hx\right)$ in each such computation.

\subsection{Details and Improvements}
We chose to focus on two different inverse problems - the denoising and deblurring problem scenarios.
In both, we chose BM3D as our Gaussian denoiser, as it provides very good results and has an efficient implementation. Both scenarios required appropriate choice of parameters and their values. An important parameter is $\beta$ - the prior weight. Empirical results show that inappropriate choice of $\beta$ leads to poor results, sometimes by several dB. Another two parameters that has big effect on the reconstruction relate to the choice of $\lambda$. This parameter is considered to be proportional to the inverse of the step size. We found that increasing $\lambda$ at each iteration gives better results then a constant $\lambda$. Thus, $\lambda$ update requires two parameters. The first is $\lambda_0$ - the initial value of $\lambda$, and the second, $\lambda_{step}$ - the value $\lambda$ is multiplied by at each iteration.
Another important parameter is the number of iterations. We chose to use a fixed number of iterations, but of course this parameter can also be learned or even estimated, similarly to what is done in \cite{romano2013improving}. 
 For a given noise peak, each parameter was tuned on a series of 8 images. Out of each original image we generated 5 degraded images, noisy for the denoising scenario, and blurred and noisy for the deblurring scenario. We tested multiple peak values in the range of 0.2 - 4. We found that the optimal $\lambda_0$ and $\beta$ have strong correlation to the peak value. On the other hand, $\lambda_{step}$ has a weak dependence on the peak, and was thus chosen independently of it.

Another improvement used, called binning, gives significant improvement in very low SNR cases. Here the image is down sampled and the algorithm is applied on the smaller sized image that has a better SNR, since we add up the photon counts of the merged pixels. Once the final result of the algorithm is obtained, we apply up-scaling by a simple linear interpolation. This technique leads to better results, and also reduces runtime as we are operating on smaller images. All the experiments reported below with binning use a 3:1 shown-scaling in each axis. Binning can only be used in the denoising scenario as down sampling doesn't commute with the blur operator.

In the denoising scenario we also tested the multiple prior P$^{\boldsymbol{4}}$IP Algorithm. We call this variation M-P$^{\boldsymbol{4}}$IP. As our second denoiser we chose to use a simplified version of \cite{sulam2014image}, which is a multi-scale denoising algorithm. Multi-scale considerations are not used in BM3D, and therefore the two algorithms joint together may form a more powerful denoising prior.

In the deblurring scenario,L-BFGS was chosen as our optimization method. In order to avoid calculating $\ln(\cdot) $ where $Hx$ is negative, we optimized the surrogate function 
\begin{equation}
f\left( x \right) = \left\{ {\begin{array}{*{20}{c}}
	{{L_\lambda }\left( x \right)\,\,\,,\,\,\,x < \varepsilon }\\
	{a{x^2} + bx + c\,\,\,,\,\,\,x \ge \varepsilon }
	\end{array}} \right.
\end{equation}
where the coefficients $a,b$ and $c$ were chosen such that this function and it's derivative coincides with $L_\lambda$ and it's derivative at $x=\epsilon$. 
as $x \to 0 $ we get that $Hx \to 0$ and $L_\lambda(x) \to \inf$, therefore choosing a small enough $\epsilon$ value, guarantees that the surrogate function will have the same minimum as $L_\lambda$ and all entries in $Hx$ are positive.

 Another technique that improves the results in both denoising and debluring states that we can apply several algorithm runs with slightly different parameters and average the final results. Of course, this comes at cost of run time. All results shown here are without this trick.

\section{Experiments}\label{section_EXPERIMENTS}
\subsection{Denoising}
We tested our algorithm for peak values 0.1, 0.2, 0.5, 1, 2 and 4.
To evaluate our algorithm we compared to BM3D with the refined inverse Anscombe transform \cite{makitalo2011optimal}. We also compared to \cite{giryes2014sparsity}, which leads to the best of our knowledge, to state of the art results. All algorithms were tested with and without binning. The results are shown in Table \ref{denoising PSNRs table}.

\begin{figure}[H]
	\centering
	\begin{subfigure}{0.3\columnwidth}
		\includegraphics[scale=0.8,clip,trim=2cm 1cm 2cm 0.8cm]{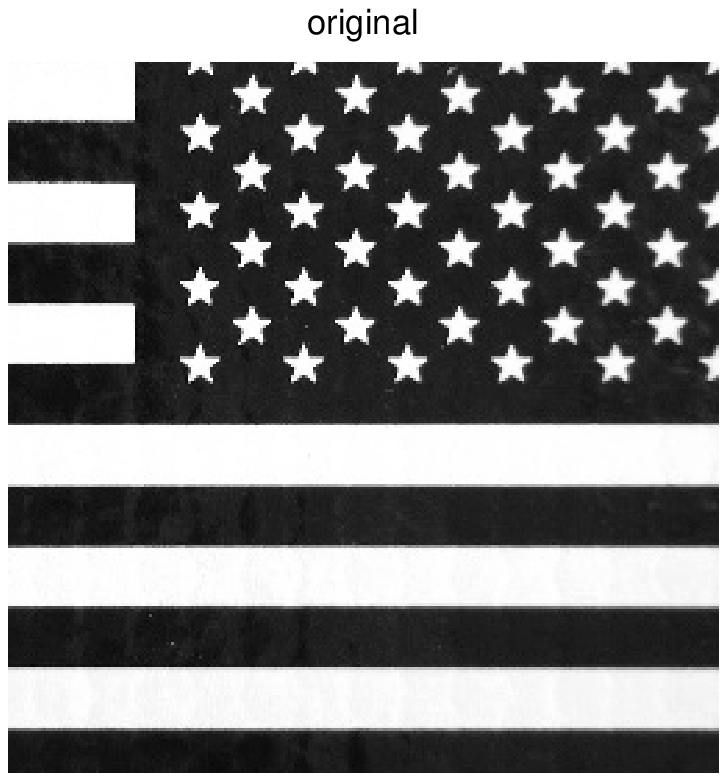}
		\caption*{original}
		\label{img::flag_original}
		
		\includegraphics[scale=0.8,clip,trim=2cm 1cm 2cm 0.8cm]{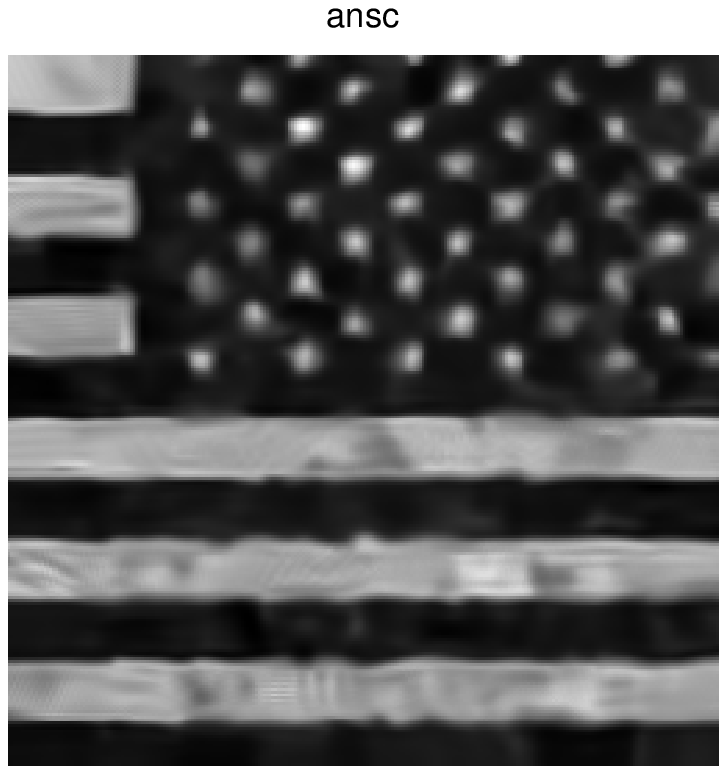}
		\caption*{Anscome+BM3D, PSNR=18.51}			
		\label{img::ansc_denoise_flag}
			
	\end{subfigure}
	\begin{subfigure}{0.3\columnwidth}	
		\includegraphics[scale=0.8,clip,trim=2cm 1cm 2cm 0.8cm]{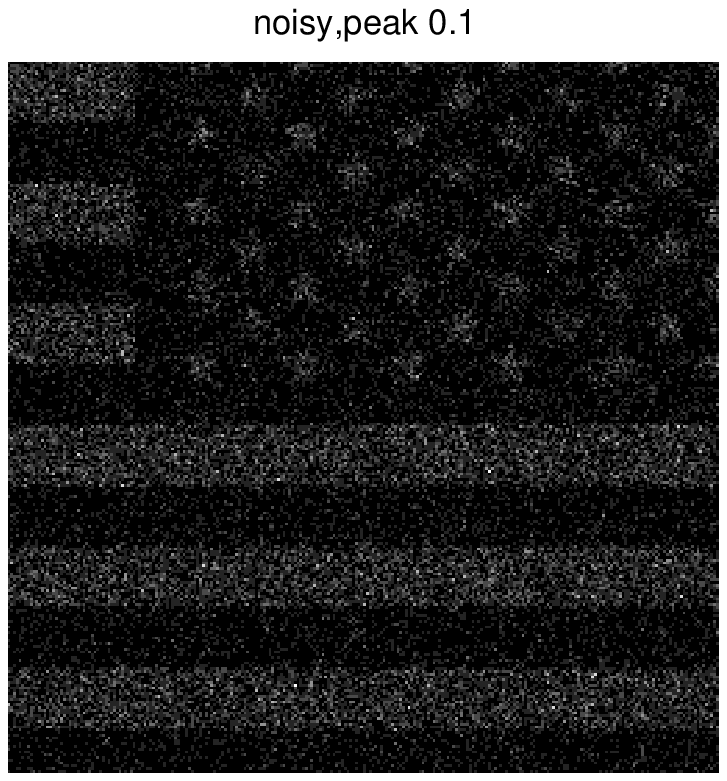}
		\caption*{noisy, peak=1}
		\label{img::flag_noisy}
		
		\includegraphics[scale=0.8,clip,trim=2cm 1cm 2cm 0.8cm]{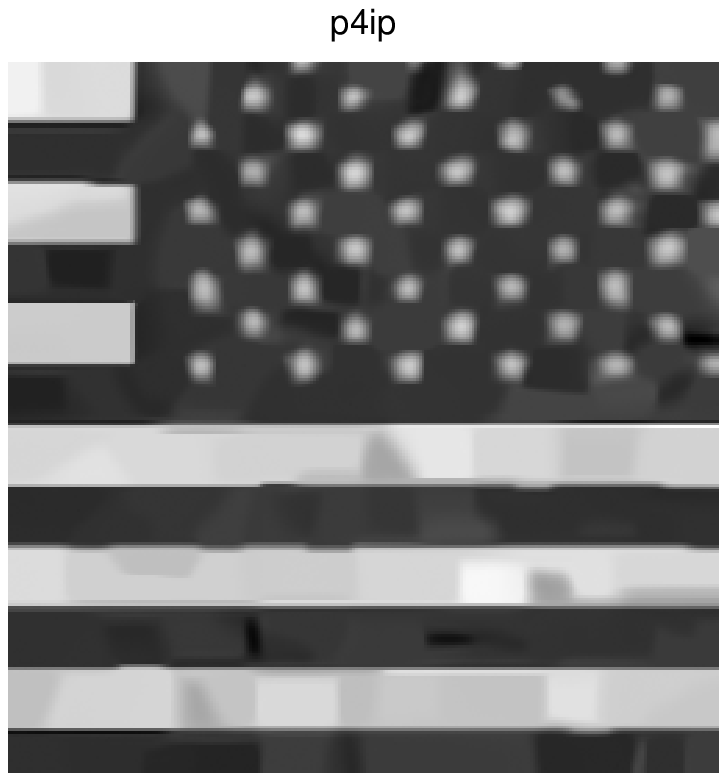}
		\caption*{P$^4$IP, PSNR=19.33}
		\label{img::p4ip_denoise_flag}
	\end{subfigure}	
	\caption{The image Flag with peak 1 - Denoising (no binning) results.}
	\label{4 images::flag denoising}
\end{figure}

\begin{figure}[H]
	\centering
	\begin{subfigure}{0.3\columnwidth}
		\includegraphics[scale=0.75,clip,trim=2cm 1cm 2cm 0.8cm]{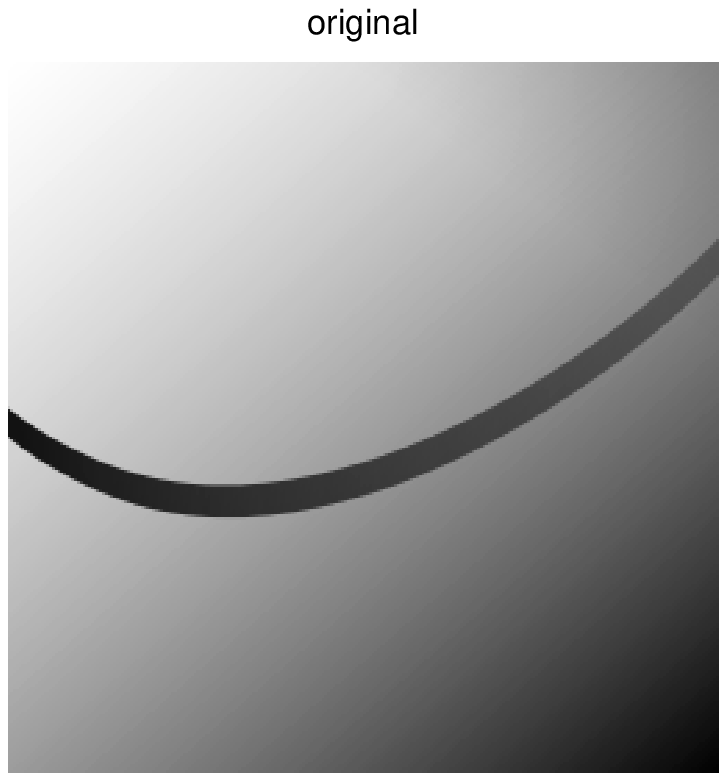}
		\caption*{original}
		\label{img::swoosh_original}
		
		\includegraphics[scale=0.75,clip,trim=2cm 1cm 2cm 0.8cm]{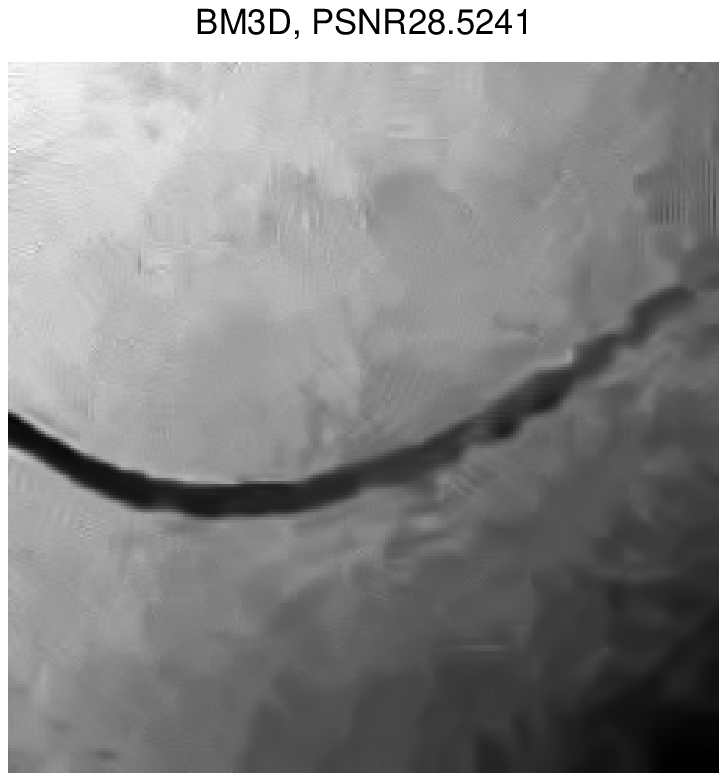}
		\caption*{Anscome+BM3D, PSNR=18.51}			
		\label{img::ansc_denoise_swoosh}
		
	\end{subfigure}
	\begin{subfigure}{0.3\columnwidth}	
		\includegraphics[scale=0.75,clip,trim=2cm 1cm 2cm 0.8cm]{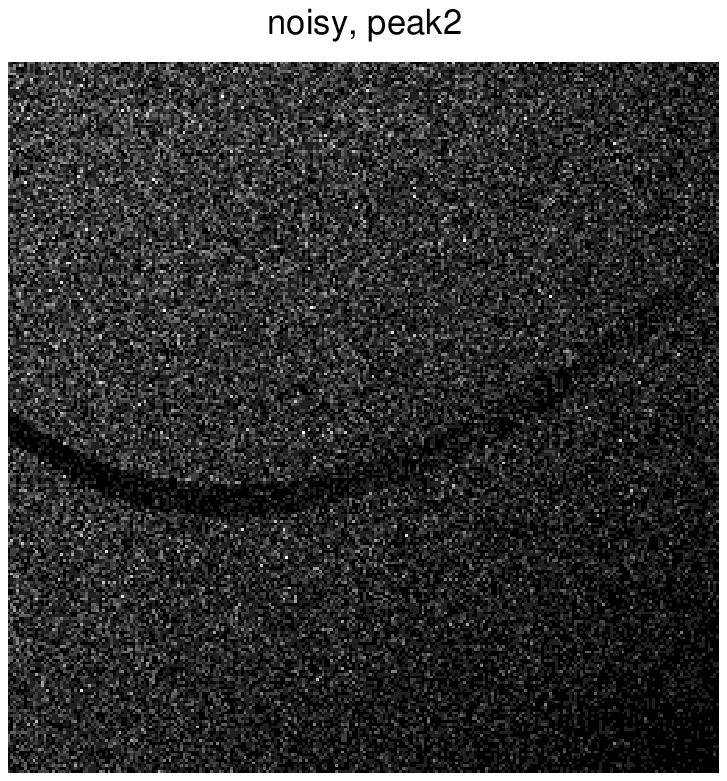}
		\caption*{noisy, peak=2}
		\label{img::swoosh_noisy}
		
		\includegraphics[scale=0.75,clip,trim=2cm 1cm 2cm 0.8cm]{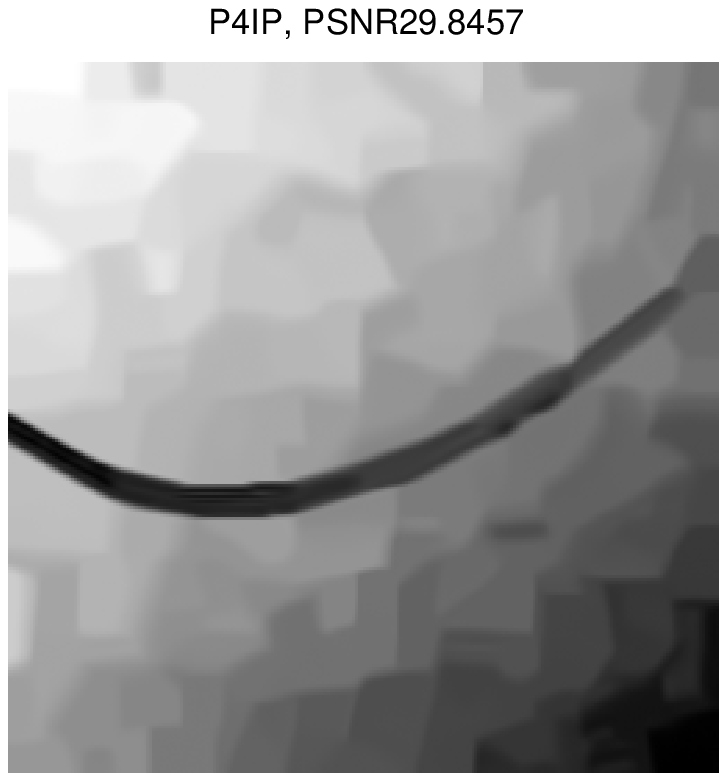}
		\caption*{P$^4$IP, PSNR=19.33}
		\label{img::p4ip_denoise_swoosh}
	\end{subfigure}	
	\caption{Peak 2 - Denoising (no binning) results}
	\label{4 images::swoosh denoising}
\end{figure}

\begin{figure}[H]
	\centering
	\begin{subfigure}{0.3\columnwidth}
		\includegraphics[scale=0.8,clip,trim=2cm 1cm 2cm 0.6cm]{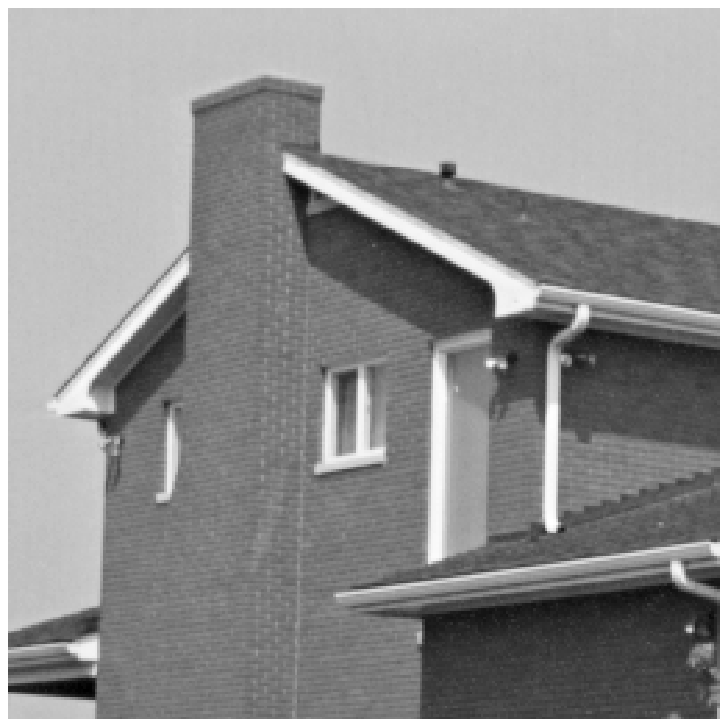}
		\caption*{original}
		\label{img::house_original}
		
		\includegraphics[scale=0.8,clip,trim=2cm 1cm 2cm 0.6cm]{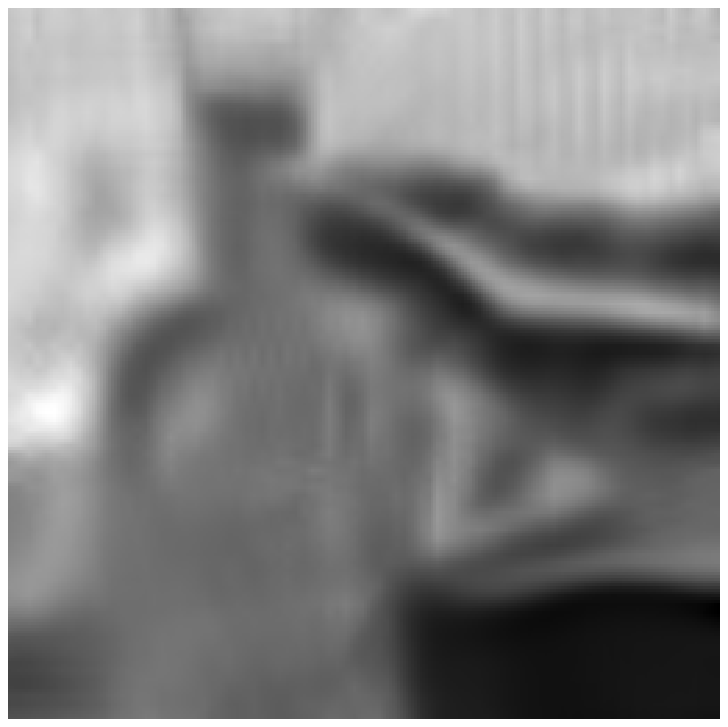}
		\caption*{Anscome+BM3D, PSNR=19.90}			
		\label{img::ansc_denoise_house}		
	\end{subfigure}
	\begin{subfigure}{0.3\columnwidth}	
		\includegraphics[scale=0.8,clip,trim=2cm 1cm 2cm 0.6cm]{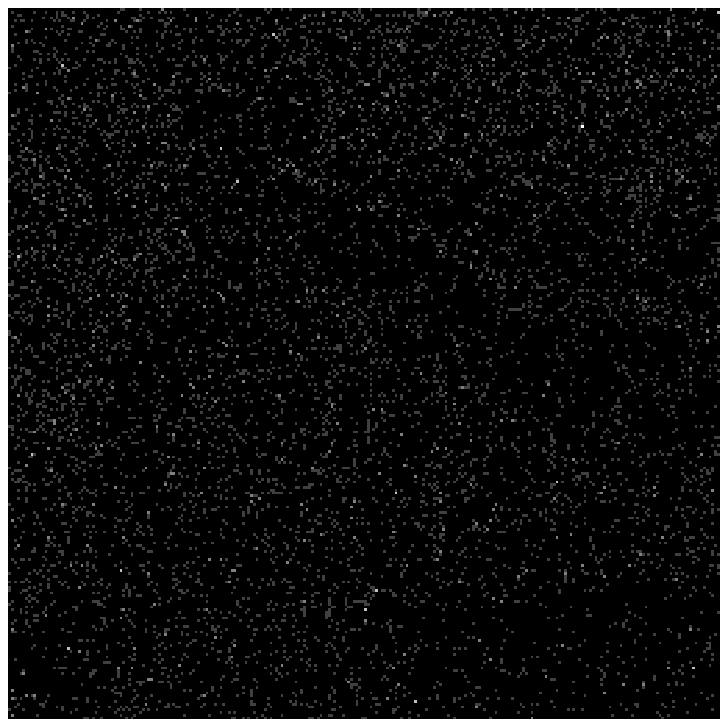}
		\caption{noisy, peak=0.2}
		\label{img::house_noisy}
		
		\includegraphics[scale=0.8,clip,trim=2cm 1cm 2cm 0.6cm]{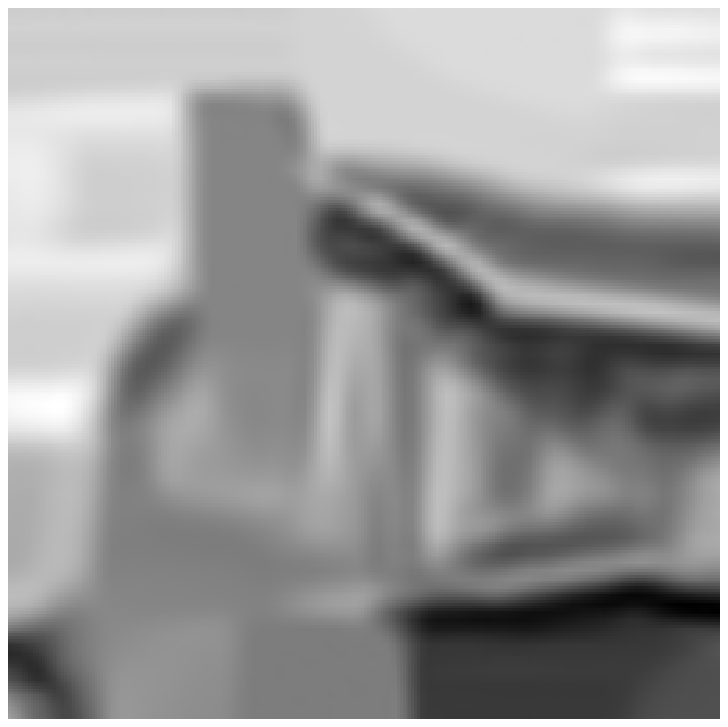}
		\caption{M-P$^4$IP, PSNR=20.43    }
		\label{img::ansc_denoise_m_p4ip}
	\end{subfigure}	
	\caption{Peak 0.2 denoising (with binning)}
	\label{4 images::house denoising}
\end{figure}

\begin{figure}[H]
	\centering
	\begin{subfigure}{0.3\columnwidth}
		\includegraphics[scale=0.8,clip,trim=2cm 1cm 2cm 0.6cm]{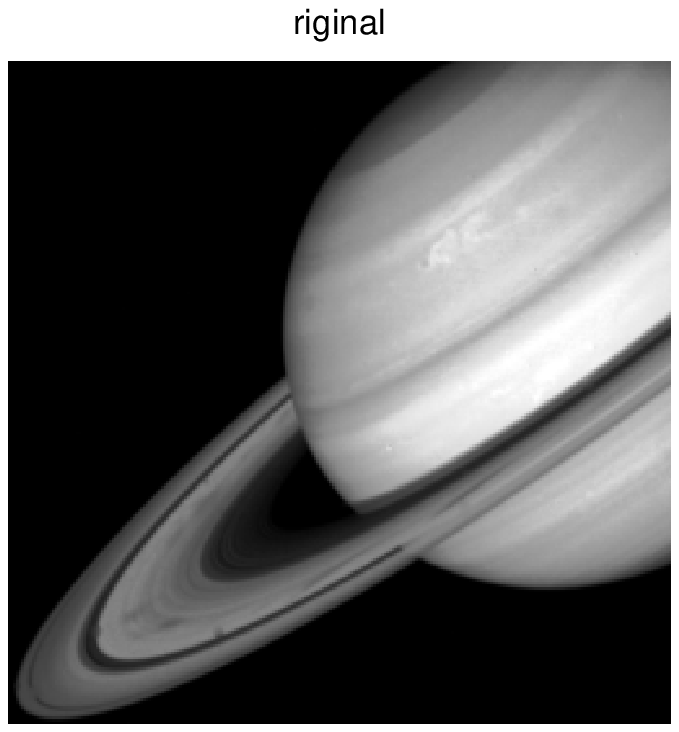}
		\caption*{original}
		\label{img::saturn_original}
		
		\includegraphics[scale=0.8,clip,trim=2cm 1cm 2cm 0.6cm]{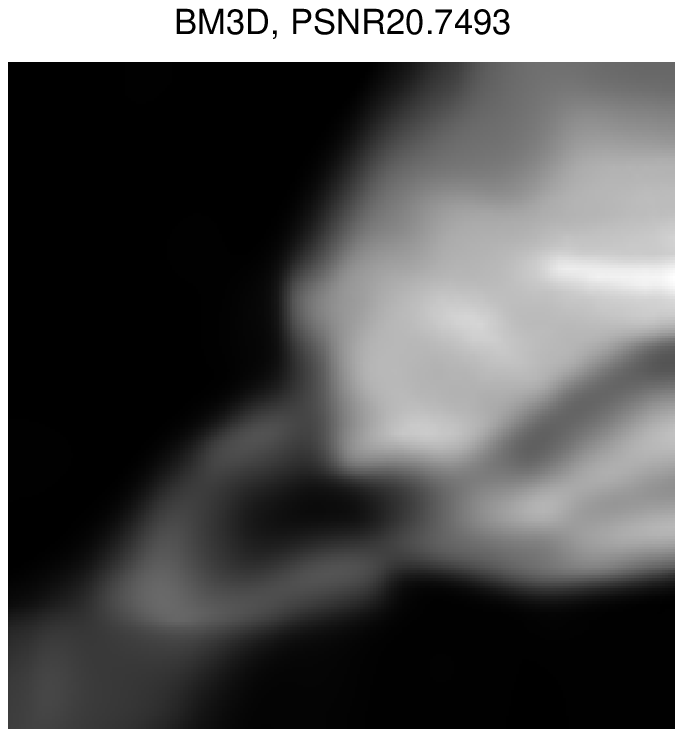}
		\caption*{Anscome+BM3D, PSNR=20.75}			
		\label{img::ansc_denoise_saturn}		
	\end{subfigure}
	\begin{subfigure}{0.3\columnwidth}	
		\includegraphics[scale=0.8,clip,trim=2cm 1cm 2cm 0.6cm]{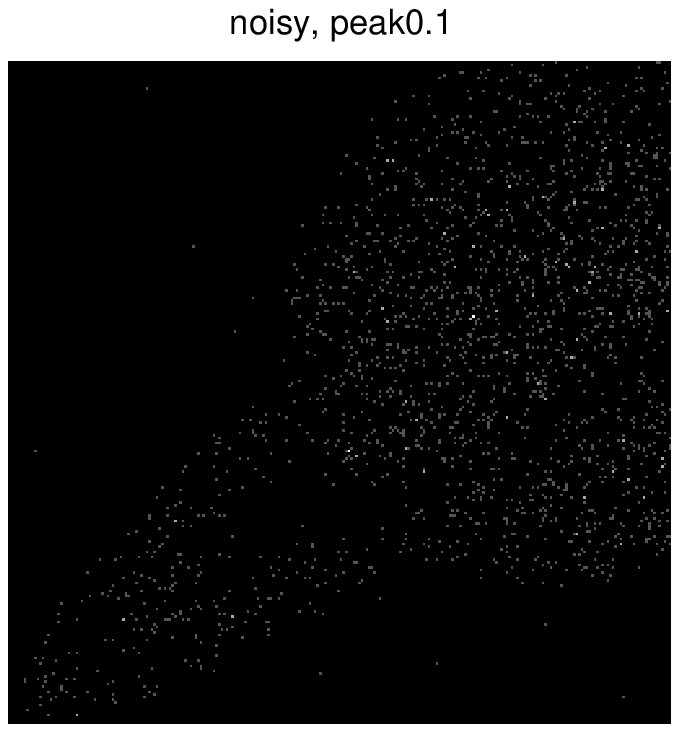}
		\caption{noisy, peak=0.2}
		\label{img::saturn_noisy}
		
		\includegraphics[scale=0.8,clip,trim=2cm 1cm 2cm 0.6cm]{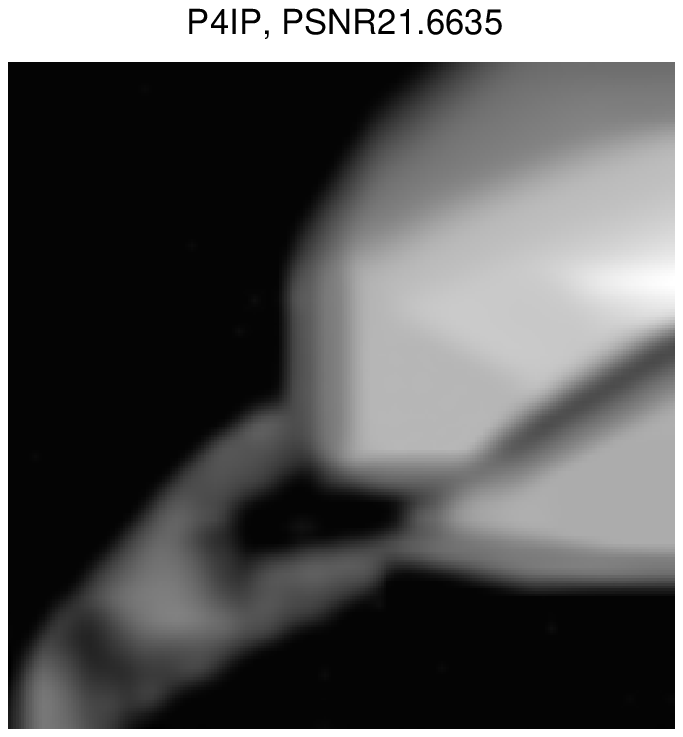}
		\caption{M-P$^4$IP, PSNR=21.66    }
		\label{img::saturn_ansc_denoise_m_p4ip}
	\end{subfigure}	
	\caption{Peak 0.2 denoising (with binning)}
	\label{4 images::saturn denoising}
\end{figure}

Figures \ref{4 images::flag denoising}-\ref{4 images::saturn denoising} show several such results. As can be seen, the propose approach competes favorably with the BM3D+Anscombe and state-of-the-art algorithms. Binning is found to be beneficial for all algorithms when dealing with low peak values. As for run-times, our algorithm takes on roughly 30 sec/image when binning is used. This should be compared to the BM3D+Anscombe that runs somewhat faster (0.5 sec/image), and the SPDA method [9] which is much slower (an average of 15-20 minutes/image).  When removing the binning, our algorithm runs for few minutes, BM3D+Anscombe runs for several seconds and SPDA runs for approximately 10 hours. These runtime evaluations were measured on an i7 with 8GB RAM laptop.

As mentioned before, to check the effectivity M-P\&P  we chose to combine BM3D with a simplified version of \cite{sulam2014image}. We noticed that it is important to find the right prior weight parameter. We only tested on a peak=0.2 scenario and the results are shown in table \ref{mult plug and play PSNRs table}. For the tested peak, the algorithm gained 0.2 dB improvement. We note that it was harder to find good parameters and therefore we believe that it is possible to improve even more.

\begin{table}[H]
	\centering
	\caption{denoising without binning PSNR values}
	\begin{tabular}{|l|c|l|l|l|l|l|l|l|l|l|l|}
		\hline
		Method & Peak  & Saturn & Flag  & Camera & House & Swoosh & Peppers & Bridge & Ridges &       & Average \\
		\hhline{============}

		BM3D    & 0.1   & 19.42 & 13.05 & 15.66 & 16.28 & 16.93 & 15.61 & 15.68 & \textbf{20.06} &	& 16.59 \\
		SPDA    &       & 17.40 & \textbf{13.35} & 14.36 & 14.84 & 15.12 & 14.28 & 14.60 & 19.86 &	& 15.48 \\
		P$^4$IP &       & \textbf{21.55} & 13.30 & \textbf{16.88} & \textbf{18.30} & \textbf{20.93} & \textbf{16.28} & \textbf{16.45} & 19.08 & & \textbf{17.85} \\

		\hline
		
		BM3D    & 0.2   & 22.02 & 14.28 & 17.35 & 18.37 & 19.95 & 17.10 & 17.09 & 21.27 &	& 18.43 \\
		SPDA    &       & 21.52 & \textbf{16.58} & 16.93 & 17.83 & 18.91 & 16.75 & 16.80 & \textbf{23.25} &	& 18.57 \\
		P$^4$IP &       & \textbf{23.05} & 14.82 & \textbf{17.82} & \textbf{19.48} & \textbf{23.34} & \textbf{17.31} & \textbf{17.54} & 21.28 & & \textbf{19.33} \\

		\hline
	
		BM3D    & 0.5   & 23.86 & 15.87 & 18.83 & 20.27 & 22.92 & 18.49 & 18.24 & 23.37 &	& 20.23 \\
		SPDA    &       & \textbf{25.50} & \textbf{19.67} & 18.90 & 20.51 & 24.21 & 18.66 & 18.46 & \textbf{27.76} &	& \textbf{21.71} \\
		P$^4$IP &       & 25.19 & 16.50 & \textbf{19.27} & \textbf{20.93} & \textbf{25.58} & \textbf{18.86} & \textbf{18.47} & 23.57 & & 21.05 \\

		\hline
		
		BM3D    & 1     & 25.89 & 18.31 & 20.37 & 22.35 & 26.07 & 19.89 & 19.22 & 26.26 &	& 21.73 \\
		SPDA    &       & 27.02 & \textbf{22.54} & 20.23 & \textbf{22.73} & 26.28 & 19.99 & 19.20 & \textbf{30.93} &	& \textbf{23.61} \\
		P$^4$IP &       & \textbf{27.05} & 19.07 & \textbf{20.54} & 22.67 & \textbf{27.79} & \textbf{20.07} & \textbf{19.31} & 26.56 & & 22.88 \\

		\hline
		
		BM3D    & 2     & 27.42 & 20.81 & \textbf{22.13} & 24.18 & 28.09 & \textbf{21.97} & \textbf{20.31} & 29.82 &	& 23.56 \\
		SPDA    &       & \textbf{29.38} & \textbf{24.92} & 21.54 & \textbf{25.09} & 29.27 & 21.23 & 20.15 & \textbf{33.40} &	& \textbf{25.62} \\
		P$^4$IP &       & 28.93 & 21.04 & 21.87 & 24.65 & \textbf{29.65} & 21.33 & 20.16 & 29.97 & & 24.70 \\
		
		\hline

		BM3D    & 4     & 29.40 & 23.04 & \textbf{23.94} & 26.04 & 30.72 & \textbf{24.07} & \textbf{21.50} & 32.39 &	& 26.39\\
		SPDA    &       & \textbf{31.04} & \textbf{26.27} & 21.90 & 26.09 & \textbf{33.20} & 22.09 & 20.55 & \textbf{36.05} &	& \textbf{27.15} \\
		P$^4$IP &       & 30.82 & 22.49 & 23.29 & \textbf{26.33} & 31.80 & 23.88 & 21.11 & 31.98 & & 26.46 \\
		\hline
		
	\end{tabular}
	\label{denoising PSNRs table}
\end{table}

\begin{table}[H]
	\centering
	\caption{denoising with binning for peak 0.2 PSNR values}
	\begin{tabular}{|l|c|l|l|l|l|l|l|l|l|l|l|}
		\hline
		Method & Peak  & Saturn & Flag  & Camera & House & Swoosh & Peppers & Bridge & Ridges &       & Average \\
		\hhline{============}
		
		BM3Dbin &    0.2 & 23.20  & 16.28 & 18.25 & 19.71 & 24.25 & 17.44 & 17.70 & 23.92 &       & 20.09 \\

		SPDAbin &       & \textbf{23.99} & \textbf{18.26} & 17.95 & 19.62 & 23.53 & \textbf{17.59} & \textbf{17.82} & \textbf{27.22} &       & \textbf{20.75} \\
		
		P$^4$IP bin &       & 23.79 & 17.26 & \textbf{18.58} & \textbf{19.96} & \textbf{24.53} & 17.44 & 17.54 & 23.94 &	& 20.38\\

	\end{tabular}
	\label{denoising with binning PSNRs table}
\end{table}

\begin{table}[H]
	\caption{multiple priors PSNR values}
	\centering	
	\begin{tabular}{|l|c|l|l|l|l|l|l|l|l|l|l|}
		\hline
		Method & Peak  & Saturn & Flag  & Camera & House & Swoosh & Peppers & Bridge & Ridges &       & Average \\
		\hhline{============}
		P$^4$IP bin   &   0.2  & 23.79 & \textbf{17.26} & \textbf{18.58} & 19.96 & 24.53 & 17.44 & 17.54 & 23.94 &	& 20.38\\
		M-P$^4$IP bin & 	  &  \textbf{24.10} & 16.77 & \textbf{18.58} & \textbf{20.02} & \textbf{24.58} & \textbf{17.63} & \textbf{17.69} & \textbf{25.38} &	& \textbf{20.59} \\
		\hline
	\end{tabular}	
	\label{mult plug and play PSNRs table}
\end{table}

\subsection{Deblurring}
\begin{figure}[H]
	\centering
	\begin{subfigure}{0.3\columnwidth}
		\includegraphics[scale=0.8,clip,trim=2cm 1cm 2cm 0.6cm]{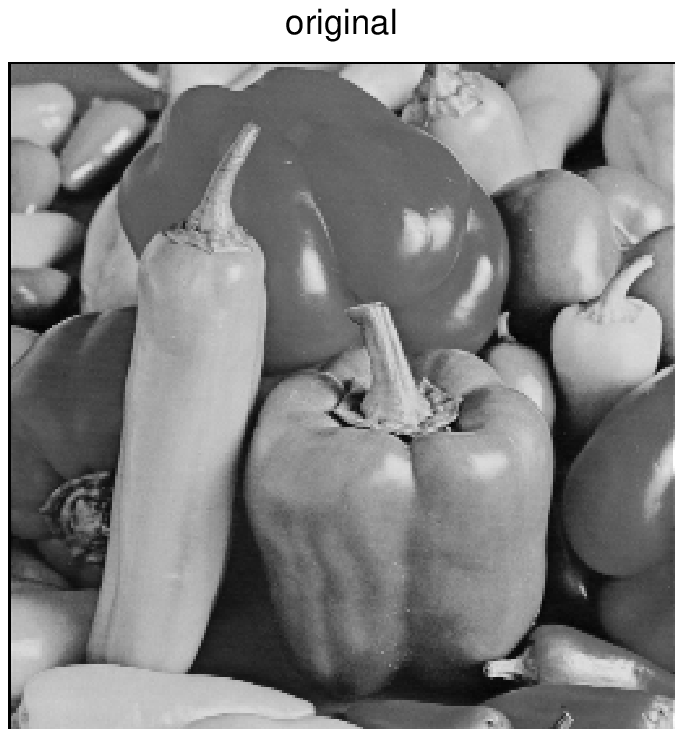}
		\caption*{original}
		\label{img::peppers_original}
		
		\includegraphics[scale=0.8,clip,trim=2cm 1cm 2cm 0.6cm]{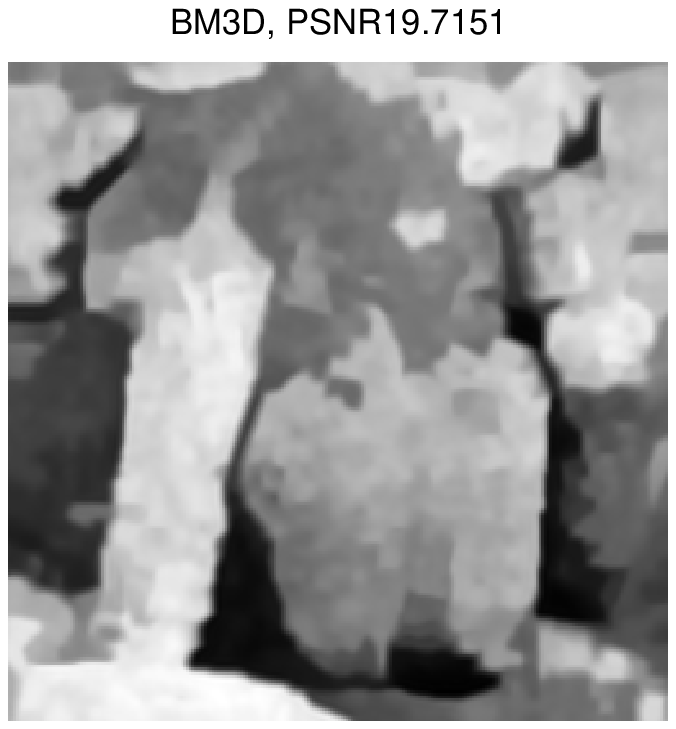}
		\caption*{Anscombe with IDD-BM3D, \\PSNR=20.65}			
		\label{img::ansc_deb_peppers}		
	\end{subfigure}
	\begin{subfigure}{0.3\columnwidth}	
		\includegraphics[scale=0.8,clip,trim=2cm 1cm 2cm 0.6cm]{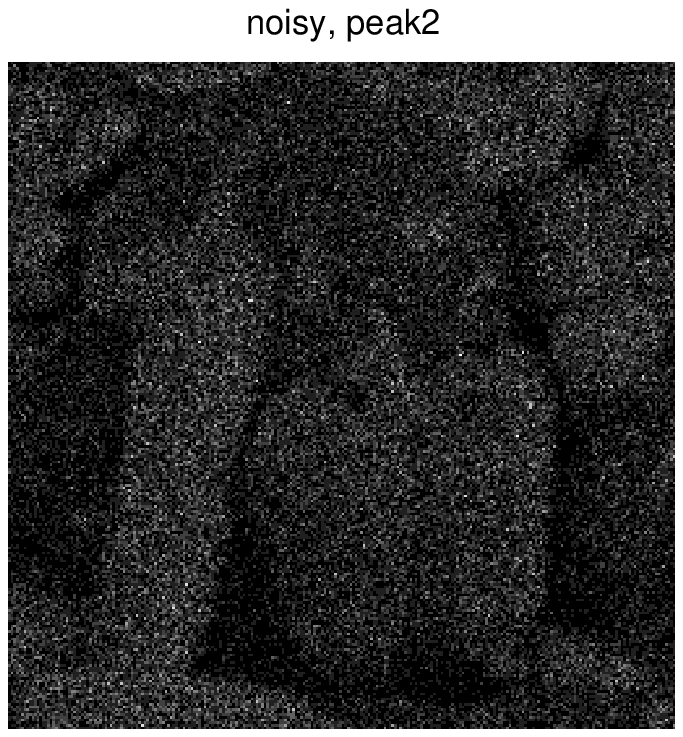}
		\caption*{noisy, peak=2}
		\label{img::peppers_noisy}
		
		\includegraphics[scale=0.8,clip,trim=2cm 1cm 2cm 0.6cm]{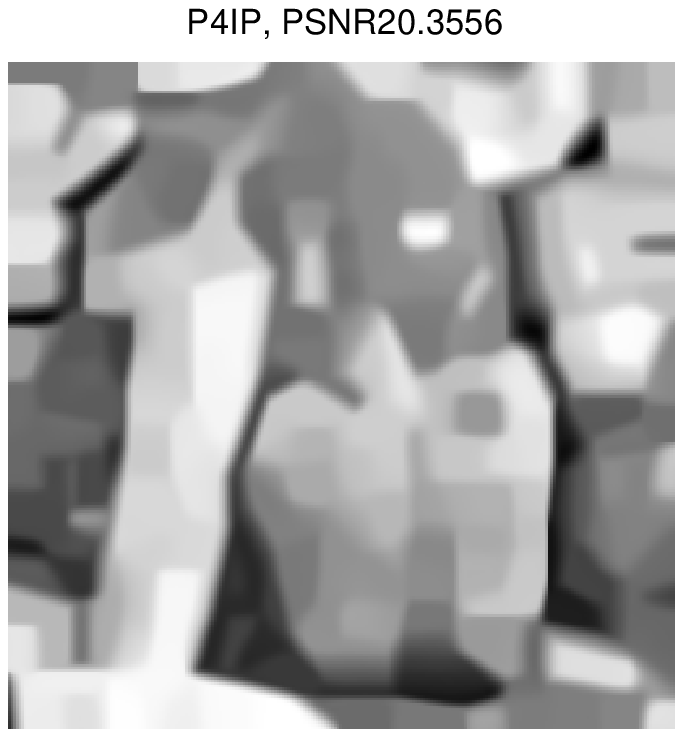}
		\caption*{P$^4$IP, \\PSNR=20.83}
		\label{img::deb_p4ip_peppers}
	\end{subfigure}	
	\caption{The image Peppers with peak 2 and blur kernel (\ref{deblurring gaussian kernel}) - deblurring results.}
	\label{4 images::peppers debluring}
\end{figure}

\begin{figure}[H]
	\centering
	\begin{subfigure}{0.3\columnwidth}
		\includegraphics[scale=0.8,clip,trim=2cm 1cm 2cm 0.6cm]{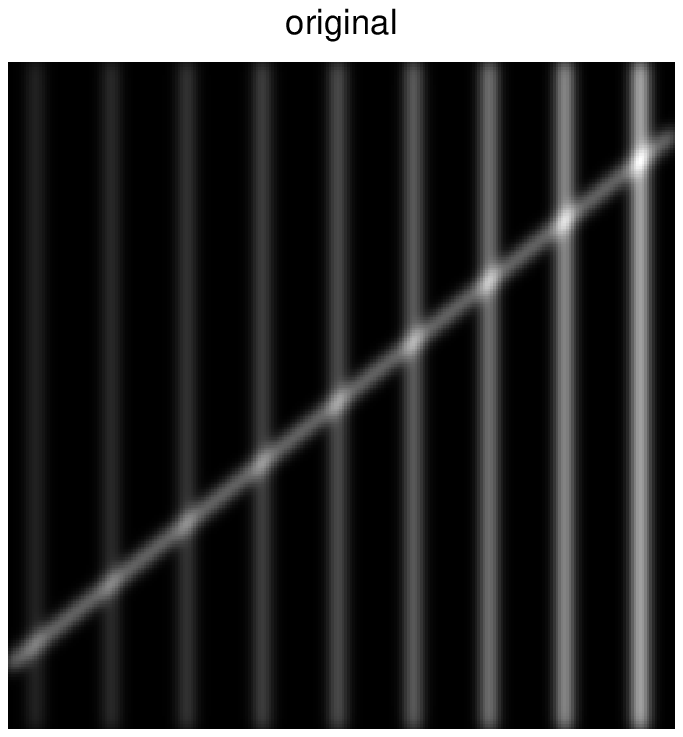}
		\caption*{original}
		\label{img::ridges_original}
		
		\includegraphics[scale=0.8,clip,trim=2cm 1cm 2cm 0.6cm]{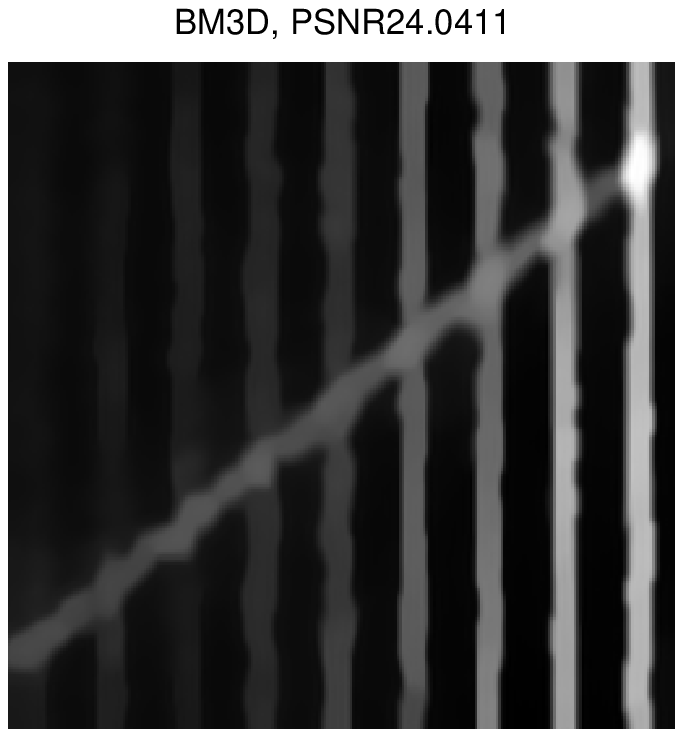}
		\caption*{Anscombe with IDD-BM3D,\\ PSNR=24.04}			
		\label{img::ansc_deb_ridges}		
	\end{subfigure}
	\begin{subfigure}{0.3\columnwidth}	
		\includegraphics[scale=0.8,clip,trim=2cm 1cm 2cm 0.6cm]{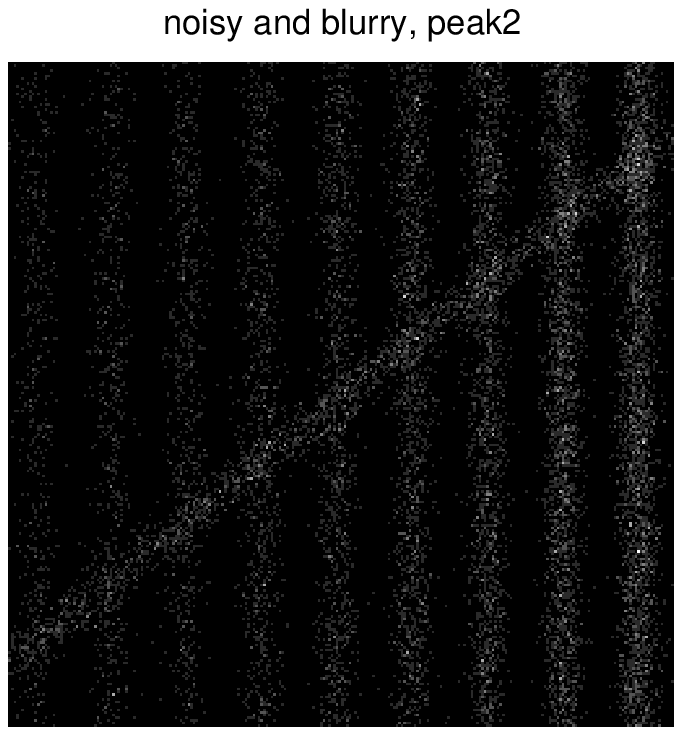}
		\caption*{noisy and blurry, peak=1}
		\label{img::ridges_noisy}
		
		\includegraphics[scale=0.8,clip,trim=2cm 1cm 2cm 0.6cm]{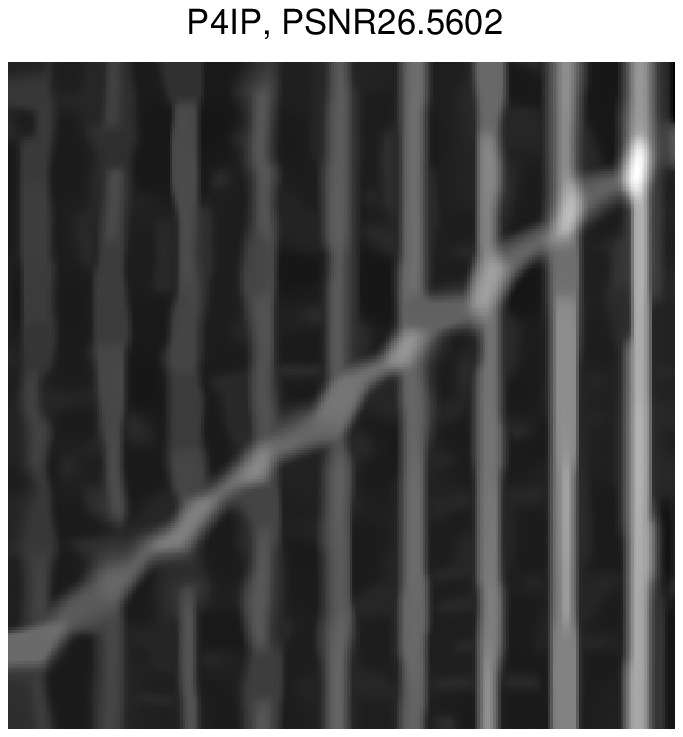}
		\caption*{P$^4$IP, \\PSNR=26.56}
		\label{img::deb_p4ip_ridges}
	\end{subfigure}	
	\caption{The image Ridges with peak 2 and blur kernel (\ref{deblurring 1/(1+x1^2+x_2^2) kernel}) - deblurring results.}
	\label{4 images::ridges debluring}
\end{figure}

\begin{figure}[H]
	\centering
	\begin{subfigure}{0.3\columnwidth}
		\includegraphics[scale=0.8,clip,trim=2cm 1cm 2cm 0.6cm]{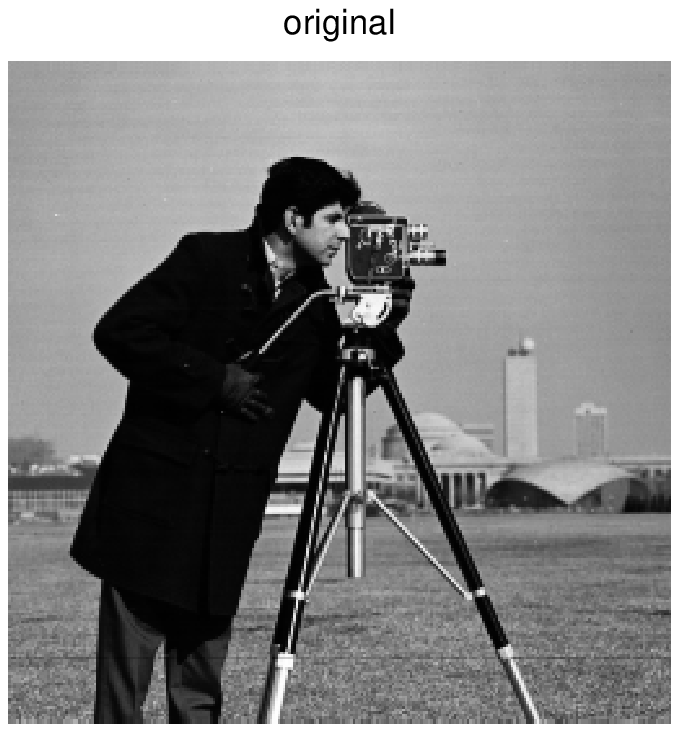}
		\caption*{original}
		\label{img::camera_original}
		
		\includegraphics[scale=0.8,clip,trim=2cm 1cm 2cm 0.6cm]{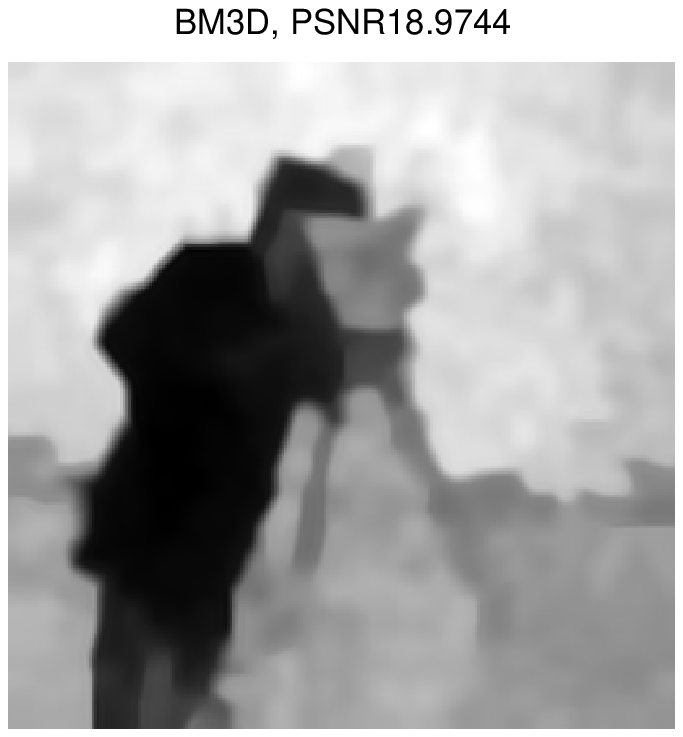}
		\caption*{Anscombe with IDD-BM3D,\\ PSNR=18.97}			
		\label{img::ansc_deb_camera}		
	\end{subfigure}
	\begin{subfigure}{0.3\columnwidth}	
		\includegraphics[scale=0.8,clip,trim=2cm 1cm 2cm 0.6cm]{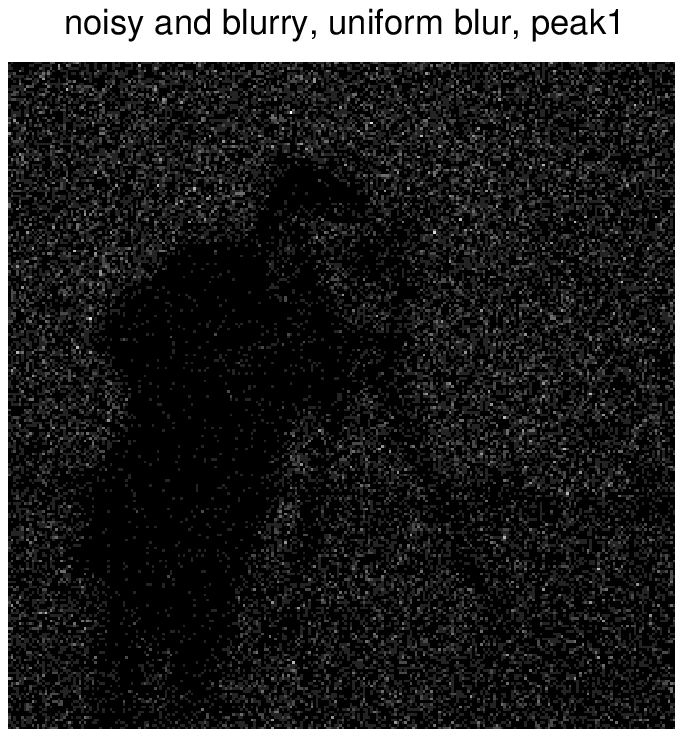}
		\caption*{noisy and blurry, peak=1}
		\label{img::camera_noisy}
		
		\includegraphics[scale=0.8,clip,trim=2cm 1cm 2cm 0.6cm]{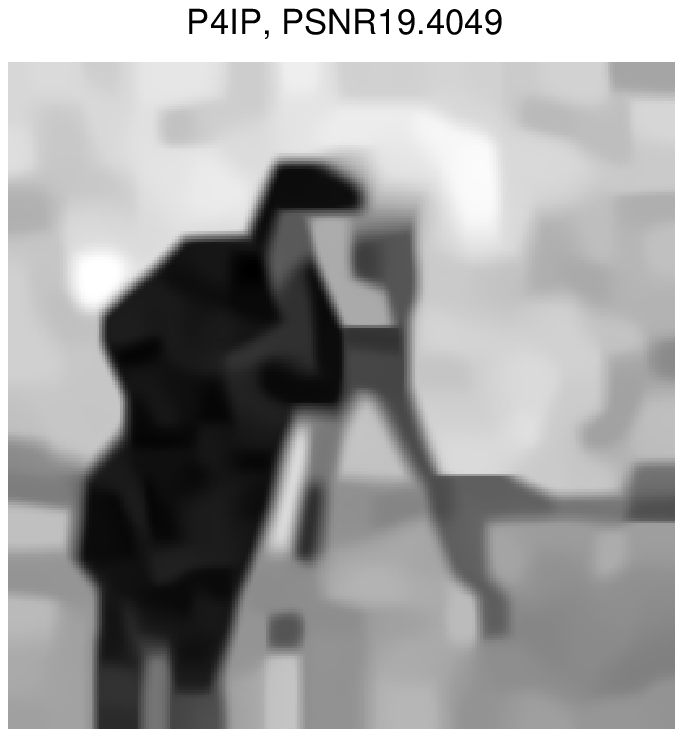}
		\caption*{P$^4$IP, \\PSNR=19.40}
		\label{img::deb_p4ip_camera}
	\end{subfigure}	
	\caption{The image Camera Man with peak 2 and blur kernel (\ref{deblurring uniform kernel}) - deblurring results.}
	\label{4 images::camera debluring}
\end{figure}

In this scenario, we tested our algorithm for the peak values 1, 2 and 4 of an image that was blurred by one of the following blur kernels:
\begin{enumerate}[(i)]
	\item
		a Gaussian kernel of size 25 by 25 with $\sigma=1.6$.
		\label{deblurring gaussian kernel}
	\item
		$\frac{1}{\left(1+x_1^2+x_2^2 \right)}$ for $x_1,x_2=-7,\dots,7$
		\label{deblurring 1/(1+x1^2+x_2^2) kernel}
	\item
		$9\times9$ uniform
		\label{deblurring uniform kernel}
\end{enumerate}
To evaluate our algorithm we compared to IDD-BM3D \cite{danielyan2012bm3d} with the refined inverse Anscombe transform \cite{makitalo2011optimal}. The results are shown in Tables \ref{deblurring PSNRs table for gaussian kernel}, \ref{deblurring PSNRs table for 1/(1+x1^2+x_2^2) kernel} and \ref{deblurring PSNRs table for uniform kernel}.
Figures \ref{4 images::peppers debluring}, \ref{4 images::ridges debluring} and \ref{4 images::camera debluring} show specific results to better assess the visual quality of the outcome. 

\begin{table}[H]
	\centering
	\caption{deblurring PSNR values for blur kernel (\ref{deblurring gaussian kernel})}
	\begin{tabular}{|l|c|l|l|l|l|l|l|l|l|l|l|}
		\hline
		Method & Peak  & Saturn & Flag  & Camera & House & Swoosh & Peppers & Bridge & Ridges &       & Average \\

		\hhline{============}

		BM3D     & 1     & 24.32 & 16.18 & 19.39 & 21.06 & \textbf{26.51} & 18.47 & 18.34 & 22.06 &     & 20.79 \\
		P$^4$IP  &       & \textbf{25.69} & \textbf{17.97} & \textbf{19.84} & \textbf{21.93} & \textbf{26.51} & \textbf{19.48} & \textbf{19.03} & \textbf{25.56} &	& \textbf{22.00}\\
		\hline
		
		BM3D  & 2     & \textbf{26.07} & 17.78 & 20.61 & 22.66 & 28.61 & 19.84 & 19.28 & 25.71 &		& 22.57 \\
		P$^4$IP  &       & 25.95 & \textbf{19.49} & \textbf{20.78} & \textbf{23.33} & \textbf{28.67} & \textbf{20.47} & \textbf{19.67} & \textbf{28.38} & & \textbf{23.34} \\
		\hline
		
		BM3D  & 4     & 28.05 & 20.25 & \textbf{21.66} & \textbf{24.69} & 30.30 & \textbf{21.25} &	\textbf{20.20} & 29.05 & 	& 24.43\\
		P$^4$IP  &       & \textbf{28.81} & \textbf{20.44} & 21.37 & 24.51 & \textbf{30.62} & 21.11 & 20.13 & \textbf{31.42} & & \textbf{24.80} \\
		\hline
		
	\end{tabular}
	\label{deblurring PSNRs table for gaussian kernel}
\end{table}

\begin{table}[H]
	\centering
	\caption{deblurring PSNR values for blur kernel (\ref{deblurring 1/(1+x1^2+x_2^2) kernel})}
	\begin{tabular}{|l|c|l|l|l|l|l|l|l|l|l|l|}
		\hline
		Method & Peak  & Saturn & Flag  & Camera & House & Swoosh & Peppers & Bridge & Ridges &       & Average \\
		
		\hhline{============}

		BM3D     & 1     & 24.36 & 15.53 & 18.99 & 20.81 & 25.83 & 18.24 & 18.20 & 21.21 & & 20.40 \\
		P$^4$IP  &       & \textbf{25.14} & \textbf{17.07} & \textbf{19.50} & \textbf{21.52} & \textbf{25.89} & \textbf{19.05} & \textbf{18.69} & \textbf{24.28} & & \textbf{21.39} \\
		\hline
		
		BM3D     & 2     & 26.02 & 16.58 & 20.01 & 22.15 & \textbf{28.33} & 19.29 & 18.98 & 24.38 & & 21.97 \\
		P$^4$IP  &       & \textbf{26.39} & \textbf{18.61} & \textbf{20.18} & \textbf{22.49} & 28.29 & \textbf{19.80} & \textbf{19.25} & \textbf{26.63} & & \textbf{22.70} \\
		\hline
		
		BM3D     & 4     & 27.64 & 19.00 & \textbf{20.84} & \textbf{23.68} & 29.45 & 20.55 & \textbf{19.71} & 27.52 & & 23.55 \\
		P$^4$IP  &       & \textbf{28.48} & \textbf{19.80} & 20.76 & 23.58 & \textbf{29.70} & \textbf{20.56} & 19.70 & \textbf{29.20} & & \textbf{23.97} \\
		\hline
		
	\end{tabular}
	\label{deblurring PSNRs table for 1/(1+x1^2+x_2^2) kernel}
\end{table}

\begin{table}[H]
	\centering
	\caption{deblurring PSNR values for blur kernel (\ref{deblurring uniform kernel})}
	\begin{tabular}{|l|c|l|l|l|l|l|l|l|l|l|l|}
		\hline
		Method & Peak  & Saturn & Flag  & Camera & House & Swoosh & Peppers & Bridge & Ridges &       & Average \\
		
		\hhline{============}

		BM3D     & 1     & 24.11 & 15.46 & 18.93 & 20.71 & \textbf{26.23} & 18.12 & 18.17 & 21.48 & & 20.40 \\
		P$^4$IP  &       & \textbf{24.36} & \textbf{17.12} & \textbf{19.49} & \textbf{21.37} & 26.03 & \textbf{19.04} & \textbf{18.64} & \textbf{23.53} & & \textbf{21.20} \\
		\hline
		
		BM3D     & 2     & \textbf{26.06} & 16.54 & 19.93 & 22.20 & \textbf{28.26} & 19.29 & 18.83 & 24.69 & & 21.97 \\
		P$^4$IP  &       & 25.62 & \textbf{18.61} & \textbf{20.11} & \textbf{22.54} & 28.17 & \textbf{19.81} & \textbf{19.19} & \textbf{25.83} & & \textbf{22.48} \\
		\hline
		
		BM3D     & 4     & 27.41 & 18.83 & 20.63 & \textbf{23.47} & 29.81 & 20.36 & 19.63 & 27.56 & & 23.46 \\
		P$^4$IP  &       & \textbf{27.97} & \textbf{19.77} & \textbf{20.66} & 23.39 & \textbf{29.93} & \textbf{20.47} & \textbf{19.71} & \textbf{29.15} & & \textbf{23.88} \\
		\hline
		
	\end{tabular}
	\label{deblurring PSNRs table for uniform kernel}
\end{table}

It is clearly shown that in this scenario P$^4$IP outperforms the Anscombe-transform framework. The runtime for a single image took about 5 minutes on an i7, 8G RAM laptop, about twice slower then Anscombe, and took 44 iterations.

\section{Conclusion and discussion}\label{section_DISCUSSION}

This work proposes a new way to integrate Gaussian denoising algorithms to Poisson noise inverse problems, by using the Plug-and-Play framework, this way taking advantage of the existing Gaussian solvers. The integretion is done by simply using the Gaussian denoiser as a "black box" as part of the overall algorithm. This work demonstrates this paradigm on two problems - image denoising and image deblurring. Numerical results show that our algorithm outperforms the Anscombe-transform framework in lower peaks, and competes favorably with it on other cases. These results could be further improved by using the proposed extension of Plug-and-Play, which enables to combine multiple Gaussian denoising algorithms.
Further work should be done in order to better tune the algorithm's parameters, similar to \cite{deledalle2010poisson}. Its is also interesting to learn more closely the relation between the Anscombe transform and our method. We have found that under certain initialization conditions, in the first step P$^4$IP does variance stabilization that is as good as Anscomb's one. It is possible that more could be said about the matter.

\begin{appendices}
	\section{Derivation of first denoising ADMM step}\label{appendix_DENOISING_ADMM_FIRST_STEP}
	In the denoising case $H=I$ and we get that $l(x)$ is given by
	\begin{equation}
		l\left( X \right) =  - {y^T}\ln \left( x \right) + {1^T}\ln \left( {\Gamma \left( {y + 1} \right)} \right) + {1^T}x.
	\end{equation}
	 The augmented Lagrangian is thus
	 \begin{equation}
		 {L_\lambda } =  - {y^T}\ln \left( {x} \right) + {1^T}x - \beta \ln \left( {P\left( v \right)} \right) + \frac{\lambda }{2}\left\| {x - v + u} \right\| - \frac{\lambda }{2}\left\| u \right\|_2^2,
	 \end{equation}
	 and the first ADMM step becomes
	 \begin{equation}
		{x^{k + 1}} = \mathop {\arg \min }\limits_x {L_\lambda }\left( {x,{v^k},{u^k}} \right) = \mathop {\arg \min }\limits_x  - {y^T}\ln \left( x \right) + {1^T}x + \frac{\lambda }{2}\left\| {x - {v^k} + {u^k}} \right\|_2^2.
	 \end{equation}
	The first step ($x$ update) is a convex and separable, implying that each entry of x can be treated separately. Furthermore, computing the elements of $x$ is easily handled leading to a closed form expression. By differentiating $L_\lambda$ by $x\left[i\right]$ and equating to 0 we get
	\begin{equation}
	- \frac{{{y[i]}}}{{{x[i]}}} + 1 + \lambda \left( {{x[i]} - {v^k}[i] + {u^k}[i]} \right) = 0.
	\label{eq::denoising_first_step_equation derivative}
	\end{equation}
	Thus, we get that
	\begin{equation}
		{x[i]} = \frac{{\left( {\lambda \left( {{v^k}[i] - {u^k}[i]} \right) - 1} \right) + \sqrt {{{\left( {\lambda \left( {{v^k}[i] - {u^k}[i]} \right) - 1} \right)}^2} + 4\lambda {y[i]}} }}{{2\lambda }}.
	\end{equation}
	As $y$ is non negative, the expression inside the square root is also non negative and causes the resulted $x$ to be non negative also. Another possible solution could have been the second root of Equation (\ref{eq::denoising_first_step_equation derivative}), but this solution is purely negative and thus uninformative.
	
\end{appendices}

\bibliography{P4IP}

\begin{thebibliography}{10}

\bibitem{anscombe1948transformation}
F.~J. Anscombe.
\newblock The transformation of poisson, binomial and negative-binomial data.
\newblock {\em Biometrika}, pages 246--254, 1948.

\bibitem{boulanger2010patch}
J.~Boulanger, C.~Kervrann, P.~Bouthemy, P.~Elbau, J.-B. Sibarita, and
  J.~Salamero.
\newblock Patch-based nonlocal functional for denoising fluorescence microscopy
  image sequences.
\newblock {\em Medical Imaging, IEEE Transactions on}, 29(2):442--454, 2010.

\bibitem{boyd2011distributed}
S.~Boyd, N.~Parikh, E.~Chu, B.~Peleato, and J.~Eckstein.
\newblock Distributed optimization and statistical learning via the alternating
  direction method of multipliers.
\newblock {\em Foundations and Trends{\textregistered} in Machine Learning},
  3(1):1--122, 2011.

\bibitem{dabov2007image}
K.~Dabov, A.~Foi, V.~Katkovnik, and K.~Egiazarian.
\newblock Image denoising by sparse 3-d transform-domain collaborative
  filtering.
\newblock {\em Image Processing, IEEE Transactions on}, 16(8):2080--2095, 2007.

\bibitem{danielyan2012bm3d}
A.~Danielyan, V.~Katkovnik, and K.~Egiazarian.
\newblock Bm3d frames and variational image deblurring.
\newblock {\em Image Processing, IEEE Transactions on}, 21(4):1715--1728, 2012.

\bibitem{deledalle2010poisson}
C.-A. Deledalle, F.~Tupin, and L.~Denis.
\newblock Poisson nl means: Unsupervised non local means for poisson noise.
\newblock In {\em Image processing (ICIP), 2010 17th IEEE international
  conference on}, pages 801--804. IEEE, 2010.

\bibitem{elad2006image}
M.~Elad and M.~Aharon.
\newblock Image denoising via sparse and redundant representations over learned
  dictionaries.
\newblock {\em Image Processing, IEEE Transactions on}, 15(12):3736--3745,
  2006.

\bibitem{fisz1955limiting}
M.~Fisz.
\newblock The limiting distribution of a function of two independent random
  variables and its statistical application.
\newblock In {\em Colloquium Mathematicae}, volume~3, pages 138--146. Institute
  of Mathematics Polish Academy of Sciences, 1955.

\bibitem{giryes2014sparsity}
R.~Giryes and M.~Elad.
\newblock Sparsity-based poisson denoising with dictionary learning.
\newblock {\em Image Processing, IEEE Transactions on}, 23(12):5057--5069,
  2014.

\bibitem{keenan2004accounting}
M.~R. Keenan and P.~G. Kotula.
\newblock Accounting for poisson noise in the multivariate analysis of tof-sims
  spectrum images.
\newblock {\em Surface and Interface Analysis}, 36(3):203--212, 2004.

\bibitem{mairal2009non}
J.~Mairal, F.~Bach, J.~Ponce, G.~Sapiro, and A.~Zisserman.
\newblock Non-local sparse models for image restoration.
\newblock In {\em Computer Vision, 2009 IEEE 12th International Conference on},
  pages 2272--2279. IEEE, 2009.

\bibitem{makitalo2011optimal}
M.~Makitalo and A.~Foi.
\newblock Optimal inversion of the anscombe transformation in low-count poisson
  image denoising.
\newblock {\em Image Processing, IEEE Transactions on}, 20(1):99--109, 2011.

\bibitem{rodrigues2008denoising}
I.~Rodrigues, J.~Sanches, and J.~Bioucas-Dias.
\newblock Denoising of medical images corrupted by poisson noise.
\newblock In {\em Image Processing, 2008. ICIP 2008. 15th IEEE International
  Conference on}, pages 1756--1759. IEEE, 2008.

\bibitem{romano2013improving}
Y.~Romano and M.~Elad.
\newblock Improving k-svd denoising by post-processing its method-noise.
\newblock In {\em ICIP}, pages 435--439, 2013.

\bibitem{salmon2014poisson}
J.~Salmon, Z.~Harmany, C.-A. Deledalle, and R.~Willett.
\newblock Poisson noise reduction with non-local pca.
\newblock {\em Journal of mathematical imaging and vision}, 48(2):279--294,
  2014.

\bibitem{schmitt2010poisson}
J.~Schmitt, J.~Starck, J.~Casandjian, J.~Fadili, and I.~Grenier.
\newblock Poisson denoising on the sphere: application to the fermi gamma ray
  space telescope.
\newblock {\em Astronomy \& Astrophysics}, 517:A26, 2010.

\bibitem{sulam2014image}
J.~Sulam, B.~Ophir, and M.~Elad.
\newblock Image denoising through multi-scale learnt dictionaries.
\newblock In {\em Image Processing (ICIP), 2014 IEEE International Conference
  on}, pages 808--812. IEEE, 2014.

\bibitem{venkatakrishnan2013plug}
S.~V. Venkatakrishnan, C.~A. Bouman, and B.~Wohlberg.
\newblock Plug-and-play priors for model based reconstruction.
\newblock 2013.

\bibitem{yu2012solving}
G.~Yu, G.~Sapiro, and S.~Mallat.
\newblock Solving inverse problems with piecewise linear estimators: from
  gaussian mixture models to structured sparsity.
\newblock {\em Image Processing, IEEE Transactions on}, 21(5):2481--2499, 2012.

\bibitem{zhang2008wavelets}
B.~Zhang, J.~M. Fadili, and J.-L. Starck.
\newblock Wavelets, ridgelets, and curvelets for poisson noise removal.
\newblock {\em Image Processing, IEEE Transactions on}, 17(7):1093--1108, 2008.

\end{thebibliography}
\bibliographystyle{abbrv}
\end{document}